\documentclass[10pt,twocolumn,letterpaper]{article}

\usepackage{cvpr}              

\usepackage[dvipsnames]{xcolor}

\definecolor{cvprblue}{rgb}{0.21,0.49,0.74}
\usepackage[pagebackref,breaklinks,colorlinks,citecolor=cvprblue]{hyperref}
\usepackage{multirow}
\usepackage{makecell}
\usepackage{textcomp,booktabs}
\usepackage{colortbl}
\usepackage{float}
\usepackage{balance}
\definecolor{mygray}{gray}{.9}

\def\paperID{9487} 
\def\confName{CVPR}
\def\confYear{2024}

\title{Generative Region-Language Pretraining for Open-Ended Object Detection}

\author{
\centerline{\textbf{Chuang Lin\textsuperscript{\rm 1}\qquad
Yi Jiang\textsuperscript{\rm 2}\thanks{Corresponding Authors}\qquad
Lizhen Qu\textsuperscript{\rm 1}\qquad
Zehuan Yuan\textsuperscript{\rm 2}\qquad
Jianfei Cai\textsuperscript{\rm 1}}\footnotemark[1]}\\
\centerline{\textsuperscript{\rm 1} Monash University \qquad
\textsuperscript{\rm 2} ByteDance Inc.
}}
\begin{document}
\maketitle
\begin{abstract}
In recent research, significant attention has been devoted to the open-vocabulary object detection task, aiming to generalize beyond the limited number of classes labeled during training and detect objects described by arbitrary category names at inference. 
Compared with conventional object detection, open vocabulary object detection largely extends the object detection categories.
However, it relies on calculating the similarity between image regions and a set of arbitrary category names with a pretrained vision-and-language model.
This implies that, despite its open-set nature, the task still needs the predefined object categories during the inference stage.
This raises the question: What if we do not have exact knowledge of object categories during inference?
In this paper, we call such a new setting as generative open-ended object detection, which is a more general and practical problem. To address it, 
we formulate object detection as a generative problem and propose a simple framework named GenerateU, which can detect dense objects and generate their names in a free-form way. 
Particularly, we employ Deformable DETR as a region proposal generator with a language model translating visual regions to object names.
To assess the free-form object detection task, we introduce an evaluation method designed to quantitatively measure the performance of generative outcomes.
Extensive experiments demonstrate strong zero-shot detection performance of our GenerateU. For example, on the LVIS dataset, our GenerateU achieves comparable results to the open-vocabulary object detection method GLIP, even though the category names are not seen by GenerateU during inference. 
Code is available at: \url{ https://github.com/FoundationVision/GenerateU}.
\end{abstract}
    
\section{Introduction}
\label{sec:intro}

\begin{figure}[t]
  \centering
   \includegraphics[width=1.1\linewidth]{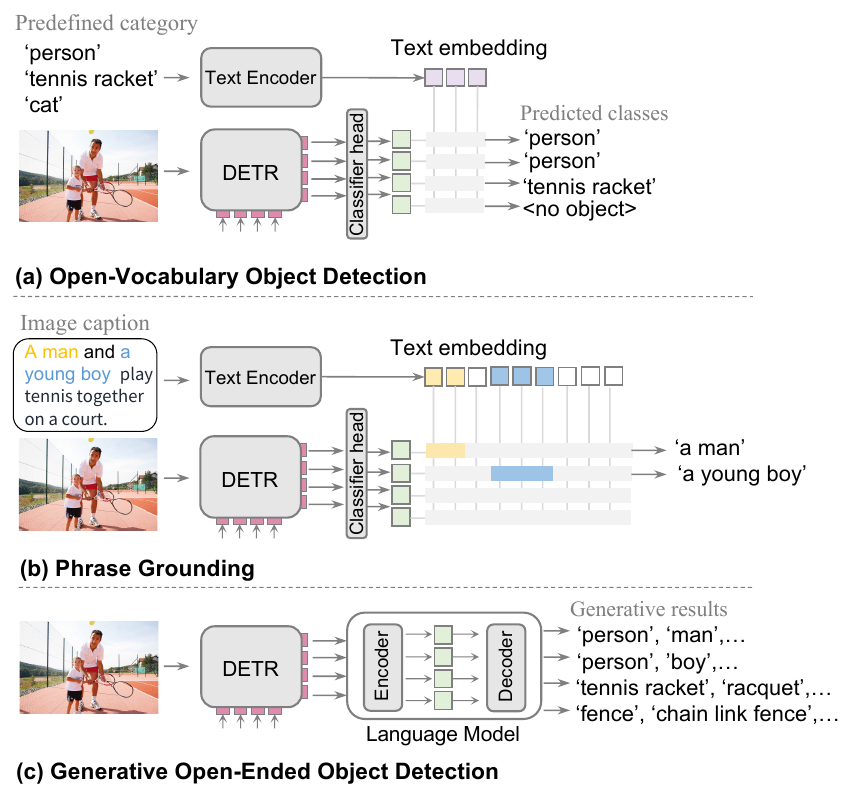}
   \caption{Comparing generative open-ended object detection with other open-set object detection tasks. Open-vocabulary object detection and phrase grounding typically require predefined categories or phrases in text prompts to align with image regions. In contrast, our introduced generative open-ended object detection 
   is a more general and practical setting 
   where categorical information is not explicitly defined. 
   Such a setting is especially meaningful for scenarios where users lack precise knowledge of object categories during inference.}
   \vspace{5mm}
   \label{fig:onecol}
\end{figure}

Object detection, a crucial task in computer vision, has made significant progress in recent years, particularly with the advent of deep learning~\cite{ren2015faster,he2017mask,cai2018cascade,carion2020end,sun2021sparse}.
The objective of this task is to locate objects in images with tight bounding boxes and recognize their respective categories.
However, 
most existing object detection algorithms are restricted to a predetermined set of object categories defined in detection datasets~\cite{chen2015microsoft,gupta2019lvis,krishna2017visual,everingham2010pascal,shao2019objects365}. For example, an object detector trained on COCO~\cite{chen2015microsoft} can well identify 80 pre-defined classes but encounters difficulty in recognizing new classes that have not been seen or defined during training.

Open-vocabulary learning has been introduced to remedy this problem and considerable efforts have been made \cite{li2021grounded,kamath2021mdetr,cai2022x,minderer2022simple,zhou2022detecting,lin2022learning, uninext}.
Open-vocabulary object detection  addresses the closed-set limitation typically via weak supervision from language, \ie, image-text pairs, or large pretrained Vision-Language
Models (VLMs), such as CLIP~\cite{radford2021learning}.
The text embeddings from the text encoder of CLIP can well align with the visual regions for the novel categories or the noun phrases in image captions, as shown in Fig.~\ref{fig:onecol}.
However, despite open-set nature, open-vocabulary object detection still requires predefined object categories during the inference stage, while in many practical scenarios, we might not have exact knowledge of test images.
%
Even if we have prior knowledge of test images, manually defining object categories
%
might introduce language ambiguities (\eg, similar object names like ``person", ``a man", ``young boy" in Fig.~\ref{fig:onecol}) or be not comprehensive enough (\eg missing ``chain link fence" in Fig.~\ref{fig:onecol}), ultimately restricting flexibility.
If we do not have prior knowledge of test images, 
to cover all possible objects that may appear in the images, the common solution to design a large and comprehensive label set that would be complex and time-consuming. The above discussions raise the research question: \textit{Can we do open-world dense object detection that does not require predefined object categories during inference?} We call this problem \textit{open-ended object detection}.


Therefore, in this paper, we formulate open-ended object detection as a generative problem and propose a simple architecture termed GenerateU to address it.
Our GenerateU contains two major components: a visual object detector to localize image regions and a language model to translate visual regions to object names.
This model is trained end-to-end, optimizing both components simultaneously.
To broaden the knowledge of GenerateU, we train it using a small set of human-annotated object-language paired data and scale up the vocabulary size with massive image-text pairs of data. Furthermore, 
due to the limitation in image captions whose nouns often fall short of accurately representing all objects in the images, we employ a pseudo-labeling method to supplement missing objects in images. 

Our major contributions are summarized as follows.
\begin{itemize}
\item {We introduce a new and more practical object detection problem: \textbf{open-ended object detection}, and formulate it as a generative problem. The reformulation avoids manual effort in predefining object categories during both training and inference, aiming at a more general and flexible architecture.}
\item {We develop a novel end-to-end learning framework, termed GenerateU, which directly generates object names in a free-form manner. GenerateU is trained with a small set of human-annotated object-language paired data and massive image-text paired data.}
\item {Compared with open-vocabulary object detection models, GenerateU achieves comparable results on zero-shot LVIS, even though it does not see object categories during inference.}

\end{itemize}

\section{Related Work}
\label{sec:related_work}
\subsection{Open-Vocabulary Object Detection}
Learning visual models from language supervision becomes popular in image recognition tasks~\cite{yu2022coca,yuan2021florence,zhai2021lit,jia2021scaling,radford2021learning,zhou2021denseclip,rao2021denseclip,huynh2021open}.
Recently, many works~\cite{kim2023contrastive,ma2023codet,yao2022detclip,yao2023detclipv2,shi2023edadet,wu2023betrayed,kaul2023multi,shi2023unified,shi2023open,kaul2023multi,wu2023cora,wang2023object,wu2023aligning,kim2023region,wang2023learning,li-etal-2023-factual} have extended the success in open-vocabulary image recognition to object detection, \ie, open-vocabulary object detection (OVOD).
As the pioneering work of OVOD, OVRCNN~\cite{zareian2021open} successfully transfers the vision and language model trained on image-text pairs to the detection framework.
To leverage the capability of CLIP~\cite{radford2021learning}, ViLD~\cite{gu2021open} and RegionCLIP~\cite{zhong2022regionclip} turn to learn the visual region feature from classification-oriented models by knowledge distillation.
OV-DETR~\cite{zang2022open} proposes a novel open-vocabulary detector built upon DETR~\cite{carion2020end} by formulating the classification as a binary matching between the input queries and the referring objects.
Different from distilling frozen vision encoder, OWL~\cite{minderer2022simple} proposes a two-step framework for contrastive image-text pre-training and end-to-end detection fine-tuning, using a standard Vision Transformer~\cite{dosovitskiy2020image}.
Detic~\cite{zhou2022detecting} proposes to extend the detector vocabulary with large-scale image classification data by applying supervision to the proposal with the largest spatial size.
To reduce the annotation cost for open-vocabulary object detection, VLDet~\cite{lin2022learning} proposes a novel framework to directly learn from image-text paired data by formulating object-language alignment as a set matching problem.

Open vocabulary object detection aims at generalization to novel categories that were not seen during the training phase. 
Existing OVOD works typically rely on calculating the similarity between image regions and a set of arbitrary category names with a pretrained vision-language model.
It remains necessary to predefine the categories during inference. 
In contrast, this work delves deeper into a new challenge - how to effectively handle scenarios where the exact categories remain unknown during the inference process.

\subsection{Multimodal Large Language Model}
Inspired by the remarkable capability of LLM~\cite{touvron2023llama,anil2023palm,brown2020language,workshop2022bloom}, multimodal foundation models~\cite{bai2023qwen,gao2023llama, li2023blip,pi2023detgpt,liu2023visual,peng2023kosmos} leverage LLM as the core intelligence and are trained with extensive image-text pairs, resulting in enhanced performance across various vision-and-language tasks such as VQA and image captioning.
BLIP-2~\cite{li2023blip} proposes a generic and efficient pre-training strategy that bootstraps vision-language pre-training from off-the-shelf frozen pre-trained image encoders and frozen large language models.
By inserting adapters into LLaMA's transformer, LLaMA-Adapter~\cite{gao2023llama} introduces only 1.2M learnable parameters, and turns a LLaMA into an instruction-following model.
LLaVA~\cite{liu2023visual} represents a novel end-to-end trained large multimodal model that combines a vision encoder and Vicuna~\cite{peng2023instruction} for general-purpose visual and language understanding. It achieves impressive chat capabilities, emulating the essence of the multimodal GPT-4.

Based on this amazing progress, subsequent studies further explore enhancing  the object localization capabilities of multimodal models.
Kosmos-2~\cite{peng2023kosmos} unlocks the grounding capability, enabling the model to provide visual answers such as bounding boxes, thereby supporting a wider array of vision-language tasks, including referring expression comprehension.
UNINEXT\cite{uninext} reformulates diverse instance perception tasks into a unified object discovery and retrieval paradigm and can flexibly perceive different types of objects by simply changing the input prompts.
DetGPT~\cite{pi2023detgpt} locates the objects of interest based on the language instructions and the LLM enables users to interact with the system, allowing for a higher level of interactivity like visual reasoning.
In contrast to these works that focus on understanding natural language expressions referring to specific objects in an image, \textit{our work aims to identify and localize all objects of interest in an image, associating each with a class name}.


\subsection{Dense Captioning}
Dense captioning aims to predict a set of descriptions across various image regions.
DenseCap~\cite{johnson2016densecap} develops a fully convolutional localization network to extract image regions, which are then described using an RNN language model.
In dense captioning, it treats each region as an individual image and tries to generate captions encompassing multiple objects and their attribute, such as ``cat riding a skateboard" and "bear is wearing a red hat".
To evaluate the generated captions, dense captioning employs the METEOR~\cite{johnson2016densecap} metric, originally designed for machine translation, which assesses the generation results from the sentence perspective.
In comparison, our proposed generative open-ended object detection focuses on describing individual objects as category names in a zero-shot setting.

Note that recently, CapDet~\cite{long2023capdet} and GRiT~\cite{wu2022grit} attempted to integrate open-vocabulary object detection and dense captioning by introducing an additional language model as a captioning head. However, for open-world object detection, they either still require a predefined category space during inference, or are limited to the close-set object detection problem.

\begin{figure*}[t]
  \centering
   \includegraphics[width=1.0\linewidth]{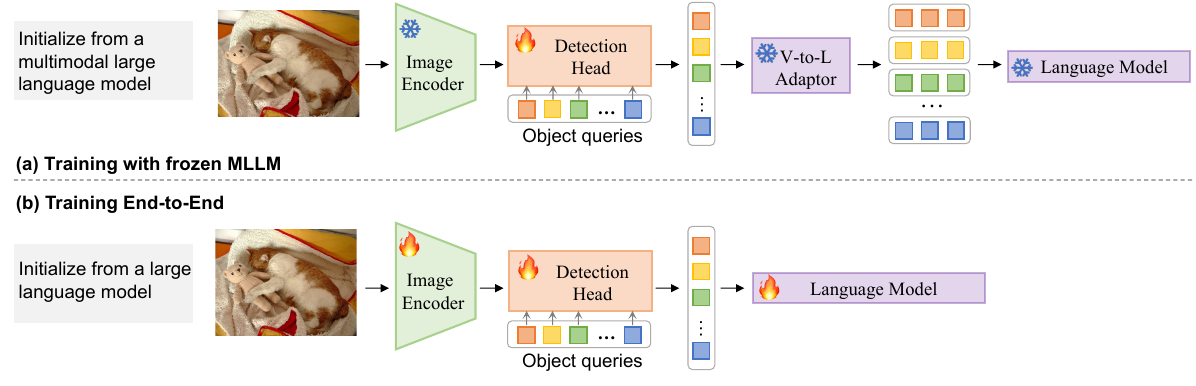}
   \caption{Overview of our proposed open-ended object detection model, GenerateU, which comprises two major components: an object detector and a language model.
   We compare two training strategies: \textit{\textbf{(Top)}} We incorporate the class-agnostic DETR (with frozen image encoder) into a pre-trained and frozen Multimodal Large Language Model (including Adaptor and Language Model), to facilitate a smooth transfer of knowledge from the language domain to object detection; \textit{\textbf{(Bottom)}} We activate the image encoder and language model as trainable components, taking an end-to-end approach to seamlessly integrate region-level understanding into the language model.}
\label{fig:fix_lm}
\end{figure*}

\section{Method}
Our goal is to build an open-world object detector capable of localizing all objects in an image and providing their corresponding category names in a free-form way.
Figure~\ref{fig:fix_lm} gives an overview of our proposed open-ended object detection model, GenerateU, which comprises two major components: an open-world object
detector and a language model. 
We describe the open-world object
detector in Sec.~\ref{sub_sec:3_1}, which consists of an image encoder and a detection head, and the language model, which can be directly borrowed from an existing Multimodal Large Language Model (including a V-to-L adaptor and a language model, Sec.~\ref{sub_sec:frozen_MLLM}) or retrained (Sec.~\ref{sub_sec:3_3}). An additional pseudo-labeling method is described in Sec.~\ref{sub_sec:3_4} for enriching the label diversity. 


\subsection{Open-World Object Detection}
\label{sub_sec:3_1}
The first crucial step is to extract accurate object regions from images, for which we develop an open-world object detector.
Although we are not constrained to any specific object detection model, 
we choose Deformable DETR~\cite{zhu2020deformable} for the following reasons. First, it alleviates heuristics when compared to other popular one or two-stage frameworks, and provides a versatile query-to-instance pipeline. Besides, Deformable DETR~\cite{zhu2020deformable} demonstrates faster convergence speed and enhanced accuracy compared to the original DETR~\cite{carion2020end}.
Deformable DETR or the original one employs Hungarian matching to learn a mapping from predicted queries to ground truth objects, followed by training the matched object query to regress to its corresponding ground truth with a combination of classification loss and bounding box regression loss.
In the context of open-set problems, we do not rely on the object class information.
Instead, we adopt an open-world detection approach, \ie class-agnostic object detector, wherein matched queries are only classified into foreground or background. Consequently, the detection process involves three losses: binary cross-entropy, generalized IoU, and L1 regression loss.

\subsection{Transferring from a Frozen Multimodal LLM}
\label{sub_sec:frozen_MLLM}
Once object candidates are localized, a natural idea is to leverage a pretrained multimodal large language model (MLLM) to reduce the training cost and offer strong zero-shot language abilities for generating object names.
Recently developed MLLMs~\cite{bai2023qwen,gao2023llama,li2023blip,pi2023detgpt,liu2023visual,peng2023kosmos} typically consist of three components: an image encoder to extract visual features; an adaptor network to bridge the modality gap and reduce the number of image embeddings, converting visual feature into token representations; a language model to generate predictions for vision-centric tasks.
As shown in Figure~\ref{fig:fix_lm} Top, we integrate a class-agnostic Deformable DETR, with a frozen image encoder to extract object queries, into an MLLM.
We keep the MLLM frozen, and the only trainable module is the detection head for learning object representations.
Besides, images are preprocessed to a resolution of $224 \times 224$ pixels during training in KOSMOS-2~\cite{peng2023kosmos}. In object detection, it is common to employ a larger input size for multi-scale feature extraction.
To handle changes in image size, we utilize linear interpolation to enlarge the original position embedding.

Despite its simplicity, our empirical study shows that such a direct open-end object detection based on pretrained MLLM results in poor performance large due to the domain shift. MLLMs were trained on large-scale image-text paired data, where they generate text descriptions given each whole image as a condition, without a fine-grained understanding of individual objects.

\subsection{Generative Region-Language Pretraining}
\label{sub_sec:3_3}
Thus, instead of employing a frozen MLLM, we propose GenerateU to directly link the open-world object detector with a language model, and activate both the image encoder and the language model as trainable components, as shown in Figure~\ref{fig:fix_lm}.
%
%
Particularly, we employ an encoder-decoder-based language model, where visual representation serves as input for the encoder, and the associated text serves as the generation target for the language decoder.
Here we only use the object queries that have been successfully matched with ground truth through Hungarian matching, ignoring the unmatched queries in the generating process.
Specifically, each matched object query is first treated as an individual region, fed to a projection function to align its dimensions with the language model.
The encoder comprises blocks with self-attention layers and feed-forward networks. The decoder, similar in structure, adds a cross-attention mechanism for interfacing with the encoder.
The decoder employs an autoregressive self-attention mechanism, limiting the model’s attention to past outputs, and its final output is processed through a dense layer with a softmax function.
We apply the language modeling loss to train the model.

\noindent\textbf{Exploring on the region-word alignment loss.}
Benefiting from the pretrained vision-language model, we incorporate a region-word alignment loss during training to aid in learning to distinguish region features. 
In detail, for an object query, we treat its corresponding word as a positive sample and consider other words in the same minibatch as negative samples. 
By calculating the similarity between the region features from class-agnostic Deformable DETR and text features from the fixed CLIP text encoder~\cite{radford2021learning}, the model is optimized by a binary cross-entropy loss.
This region-word alignment loss is jointly trained with the object detection losses and the language modeling loss.

\subsection{Enrich Label Diversity}
\label{sub_sec:3_4}
\begin{figure*}
  \centering
   \includegraphics[width=0.8\linewidth]{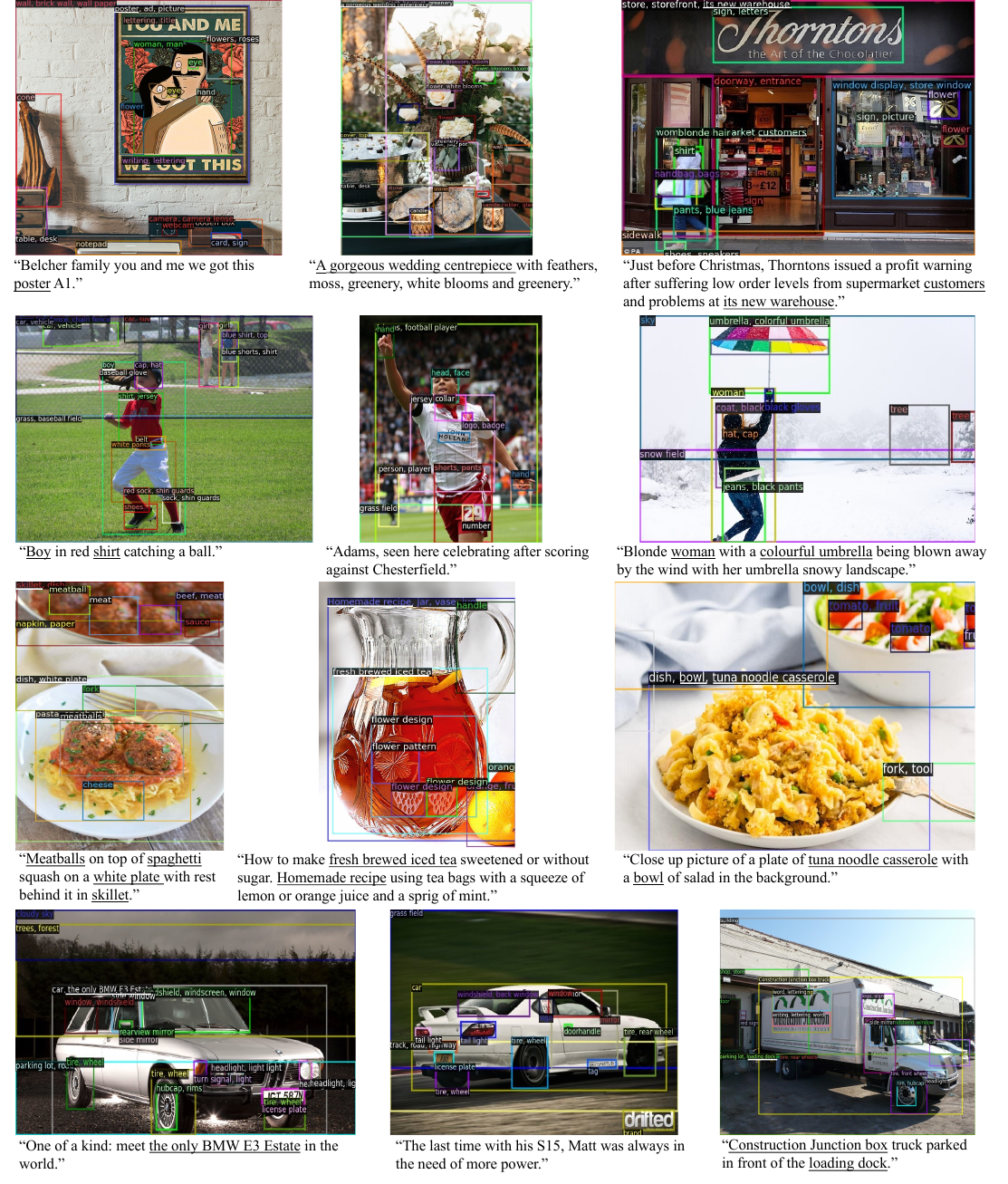}
   \caption{Selected pseudo-label examples highlight the generation of bounding boxes covering nearly all objects in the images. A varied set of descriptive labels showcases the model's capability in producing diverse and linguistically rich vocabulary. A white underline in a figure indicates that the pseudo label comes from the noun phrase (black underline) in the caption.}
   \label{fig:grit_vis}
\end{figure*}

Considerable efforts have been invested in collecting detection data that is both semantically rich and voluminous. 
However, human annotation proves to be both costly and resource-restrictive. 
Previous works, such as Kosmos-2~\cite{peng2023kosmos} and GLIP~\cite{li2021grounded}, leverage image-text pairs to expand the vocabulary scale.
They use pretrained grounding models to generate bounding boxes aligned with noun phrases from captions. These boxes, in turn, serve as pseudo detection labels for training the model.
However, given that captions may not comprehensively describe every object in an image, the count of nouns in the caption is typically much lower than the actual number of objects present in the corresponding image. 
Consequently, the quantity of pseudo-labels generated is constrained.
To address this, we use GenerateU pretrained on the available labels to generate pseudo labels as a supplement for missing objects in images.
Furthermore, we observed that beam search in language models naturally generates synonyms, thereby providing diverse object labels.
Figure~\ref{fig:grit_vis} provides pseudo-label examples, illustrating the generation of boxes covering nearly all objects in the image, along with a diverse set of rich vocabulary labels.

\section{Experiments}
\subsection{Datasets}
Our model is trained with hybrid supervision from two types of data: detection data and image-text data.
For detection data, we use Visual Genome (VG)~\cite{krishna2017visual} which contains 77,398 images for training.
Different from previous works~\cite{long2023capdet,wu2022grit}, we do not use the region captioning annotations of VG, since they may contain descriptions that encompass multiple objects or even describe actions.
Our focus is directed towards the object detection task, and thus we leverage the object annotations of VG which annotate the same bounding box regions with concise and specific object names, covering a wide range of objects.
For pretraining, we use about 5 million image-text pairs GRIT~\cite{peng2023kosmos} from COYO700M~\cite{kakaobrain2022coyo-700m} with pseudo grounding labels generated by KOSMOS-2~\cite{peng2023kosmos} using GLIP~\cite{li2021grounded}.

\setlength\tabcolsep{7pt}
\begin{table*}
  \centering
  \begin{tabular}{@{}lcccccccc@{}}
    \toprule
    \multirow{2}*{Method} & \multirow{2}*{Backbone} & \multirow{2}*{Pre-Train Data} & \multirow{2}*{Open-ended} & \multirow{2}*{Open-set} & \multicolumn{4}{c}{LVIS}\\
    & & & & & APr & APc & APf & AP \\
    \midrule
    MDETR~\cite{kamath2021mdetr} & RN101 & LVIS & - & - & 20.9 & 24.9 & 24.3 & 24.2\\
    MaskRCNN~\cite{he2017mask} & RN101 & LVIS & - & - & 26.3 & 34.0 & 33.9 & 33.3\\
    Deformable DETR~\cite{zhu2020deformable} & Swin-T & LVIS  & - & - & 24.2 & 36.0 & 38.2 & 36.0\\
    \midrule
    GLIP (C)~\cite{li2021grounded} & Swin-T & O365, GoldG & - & \checkmark &17.7 & 19.5 & \textbf{31.0} & 24.9\\
    GLIP~\cite{li2021grounded} & Swin-T & O365, GoldG, CAP4M & - & \checkmark & \textbf{20.8} & 21.4 & \textbf{31.0} & 26.0\\
    GenerateU & Swin-T & VG & \checkmark & \checkmark & 17.4 & 22.4 & 29.6 & 25.4 \\
    GenerateU & Swin-T & VG, GRIT5M & \checkmark & \checkmark & 20.0 & \textbf{24.9} & 29.8 & \textbf{26.8} \\
    \midrule
    GenerateU & Swin-L & VG, GRIT5M & \checkmark & \checkmark & \textbf{22.3} & \textbf{25.2} & \textbf{31.4} & \textbf{27.9} \\
    \bottomrule
  \end{tabular}
  \caption{Zero-shot domain transfer to LVIS. Our method achieves comparable performance to prior models in the close-ended setting, without requiring access to category names during inference.}
  \label{tab:main_results}
\end{table*}

\subsection{Evaluation protocol and metrics}
Addressing an open-ended problem presents challenges in evaluating performance. 
For instance, when annotating a laptop, the human-annotated ground truth category may be ``laptop", ``computer", ``PC", ``Microcomputer", \etc .
To assess the effectiveness of the open-end generative results, we propose two settings:
(a) We calculate the similarity score between the generated and annotated category names with a fixed pretrained text encoder.
Notice that this text encoder is not utilized by the generative detection model; rather, it is only used for quantitative performance measurement.
Any text encoder capable of computing similarity between two sentences can be adopted for this purpose. In this study, we select the popular CLIP~\cite{radford2021learning} text encoder and BERT-large~\cite{devlin2018bert} model.
In this way, we can map the generative results to predefined categories in a dataset for evaluation.
(b) We use METEOR~\cite{banerjee2005meteor} to assess the generated text quality, which is the same as the evaluation in dense captioning~\cite{johnson2016densecap}.
METEOR is widely used in NLP for evaluating machine translations.
Following \cite{johnson2016densecap}, we measure the mean Average Precision (AP) for both localization, using Intersection over Union (IoU) thresholds of .3, .4, .5, .6, and .7, and language accuracy, applying METEOR score thresholds at 0, .05, .1, .15, .2, and .25.

Following GLIP~\cite{li2021grounded}, we evaluate our GenerateU primarily on LVIS~\cite{gupta2019lvis} which contains 1203 categories and report APs for rare categories ($AP_r$), common categories ($AP_c$), frequency categories ($AP_f$) and all categories ($mAP$), respectively. Specifically, 
following GLIP~\cite{li2021grounded} and MDETR~\cite{kamath2021mdetr}, we evaluate on the 5k minival subset and report the zero-shot fixed AP~\cite{dave2021evaluating} for a fair comparison.

\subsection{Implementation Details}
We attempt two different backbones, Swin-Tiny and Swin-Large~\cite{liu2021swin} as the visual encoder.
The Deformable DETR architecture follows \cite{zhu2020deformable} with 6 encoder layers and 6 decoder layers.
The number of object queries $N$ is set to 300.
We adopt FlanT5-base~\cite{chung2022scaling} as the language model and initialize the weights from it.
The optimizer is AdamW~\cite{loshchilov2017decoupled} with a weight decay of 0.05.
The learning rate is set to $2 \times 10 ^{-4}$ for the parameters of the visual backbone and the detection head, and $3 \times 10 ^{-4}$ for the language model. 
For the frozen MLLM approach in Sec~\ref{sub_sec:frozen_MLLM}, we explore the state-of-the-art model KOSMOS-2~\cite{peng2023kosmos} which was trained with with the multimodal corpora.
We fix all parameters and set the learning rate to $2 \times 10 ^{-4}$ for the detection head.
Our model is trained on 16 A100 GPUs.

\setlength\tabcolsep{11pt}
\begin{table*}
\centering
  \begin{tabular}{@{}lcccccccc@{}}
    \toprule
    \multirow{2}*{Methods} &\multirow{2}*{\makecell[c]{Image Encoder}} & \multirow{2}*{\makecell[c]{Detection Head}} & \multirow{2}*{\makecell[c]{Language Model}} & \multicolumn{4}{c}{LVIS}\\
    & & & & APr & APc & APf & AP \\
    \midrule
    \multirow{2}*{Init from MLLM}& \textcolor{CornflowerBlue}{Freeze} & \textcolor{Bittersweet}{Train} & \textcolor{CornflowerBlue}{Freeze}\textsuperscript{\textcolor{CornflowerBlue}{$\star$}} & 5.3  & 10.2 & 19.7 & 14.3\\
    & \textcolor{Bittersweet}{Train} & \textcolor{Bittersweet}{Train} & \textcolor{CornflowerBlue}{Freeze}\textsuperscript{\textcolor{CornflowerBlue}{$\star$}} & 10.9 & 17.2 & 28.1 &21.9\\
    \midrule
    \multirow{2}*{Init from LLM}& \textcolor{Bittersweet}{Train} & \textcolor{Bittersweet}{Train} & \textcolor{CornflowerBlue}{Freeze} & 11.2  & 20.1 & 27.4 & 22.9\\
    & \textcolor{Bittersweet}{Train} & \textcolor{Bittersweet}{Train} & \textcolor{Bittersweet}{Train} & 20.0 & 24.9 & 29.8 & 26.8\\
    \bottomrule
  \end{tabular}
  \caption{Comparison of our GenerateU with two different training strategies on a pretrained LLM on zero-shot LVIS. 
  The results highlight generative object detection benefits from end-to-end training. Here, $\star$ refers to freezing both V-to-L adaptor and language model.} 
  \label{tab:freeze}
\end{table*}

\subsection{Generative Open-Ended Detection Results}
We assess the model's capacity to identify diverse objects in a zero-shot setting on LVIS, presenting the results in Table~\ref{tab:main_results}, where rows 1-3 include methods utilizing LVIS as pre-training data, representing fully supervised models.
Remarkably, we observe that even with only VG as training data, our model demonstrates commendable performance on zero-shot LVIS. 
This result 
suggests that predefined class names might not be necessary for open-world object detection during inference, especially when models are trained with extensive visual concepts. Furthermore, introducing additional training data from cost-effective image-text pairs consistently improves performance. 
For instance, when GenerateU trained with additional image-text pair data (GRIT~\cite{peng2023kosmos}), we observe a substantial performance gain of 3.6 on $AP_{r}$ and 1.4 on $AP$.
In summary, the semantic richness derived from image-text data significantly enhances the model's ability to recognize rare objects. Notably, our model achieves comparable performance to GLIP under the same training data size but without requiring access to category names during inference.
Moreover, we enhanced our model by integrating a larger backbone, \ie Swin-Large, confirming that our method can scale effectively with increased model complexity.

\textbf{Transfer to other datasets.}
Our generative open-ended object detector can transfer to any dataset without any modifications. In contrast to open vocabulary object detection methods, we do not need to switch the classifier to the specific category text embeddings. We evaluate the transferability of the proposed GenerateU on COCO~\cite{chen2015microsoft} and Objects365~\cite{shao2019objects365} validation sets, as shown in Table~\ref{tab:coco_obj365}. The results demonstrate the effectiveness of our method and underscore the importance of scaling up the training data.

\setlength\tabcolsep{10pt}
\begin{table}
  \begin{tabular}{@{}lccc@{}}
    \toprule
    Methods & Pre-Train Data  & COCO & Objects365 \\
    \midrule
    Supervised & - & 46.5 & 25.6\\
    GenerateU & VG &33.0 &10.1 \\
    GenerateU & VG, GRIT &33.6 &10.5 \\
    \bottomrule
  \end{tabular}
  \caption{We evaluate our GenerateU on COCO and Objects365 validation set in a zero-shot setting. GenerateU can transfer to various downstream datasets without requiring any modifications. }
  \label{tab:coco_obj365}
\end{table}

\setlength\tabcolsep{12pt}
\begin{table}
  \begin{tabular}{@{}lccccc@{}}
    \toprule
    Metrics & APr & APc & APf & AP \\
    \midrule
    CLIP~\cite{radford2021learning} & 20.0 & 24.9 & 29.8 & 26.8\\
    BERT~\cite{devlin2018bert} & 12.5 & 19.9 & 25.1 & 21.7 \\
    METEOR~\cite{banerjee2005meteor} & - & - & - & 9.2\\
    \bottomrule
  \end{tabular}
  \caption{Effect of different evaluation metrics on zero-shot LVIS.}
  \label{tab:metrics}
\end{table}

\subsection{Ablation Study}
\label{ab_study}
\textbf{Comparison with transferring from a Frozen MLLM.}
In Table~\ref{tab:freeze}, we evaluate the approach discussed in Sec~\ref{sub_sec:frozen_MLLM}, \ie transferring a frozen MLLM to the generative object detection task.
Following the state-of-the-art MLLM model KOSMOS-2~\cite{peng2023kosmos}, we maintain its parameters fixed and only train the detection head.
However, we observe that such direct transfer of the MLLM to object detection leads to poor performance in zero-shot LVIS.
We attribute this result to the pretrained image encoder, which is trained on images of size $224 \times 224$, which is insufficient for the object detection task. 
Besides, the MLLM originally supervised by image-level annotations such as image captions, lacks understanding in both object localization and fine-grained object categories.
We also experiment with making the image encoder trainable while keeping the language model frozen.
This adjustment significantly improves the results. We further fine-tune the entire model, initializing the Large Language Model (LLM) from FlanT5-base~\cite{chung2022scaling}, which achieves the best performance. The above analysis underscores the importance of end-to-end training in the open-ended object detection task.

\begin{figure}
  \centering
    \includegraphics[width=1.0\linewidth]{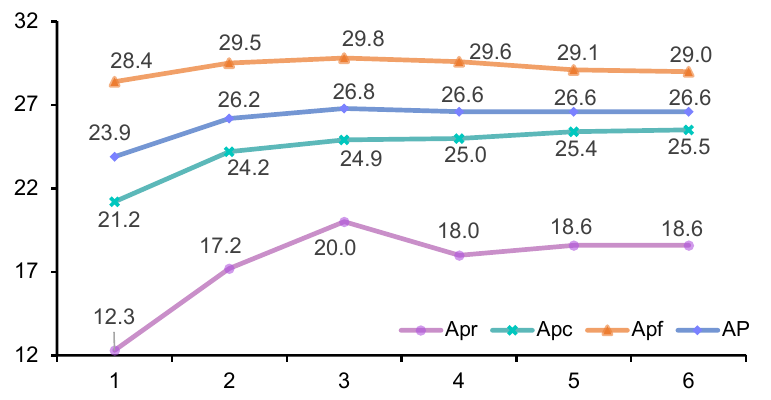}
    \caption{Effect of beam size. Beam search plays a crucial role in generating rare object names, effectively addressing the long-tail problem.}
    \label{fig:beam_search}
\end{figure}

\begin{figure*}
  \centering
  \includegraphics[width=0.9\linewidth]{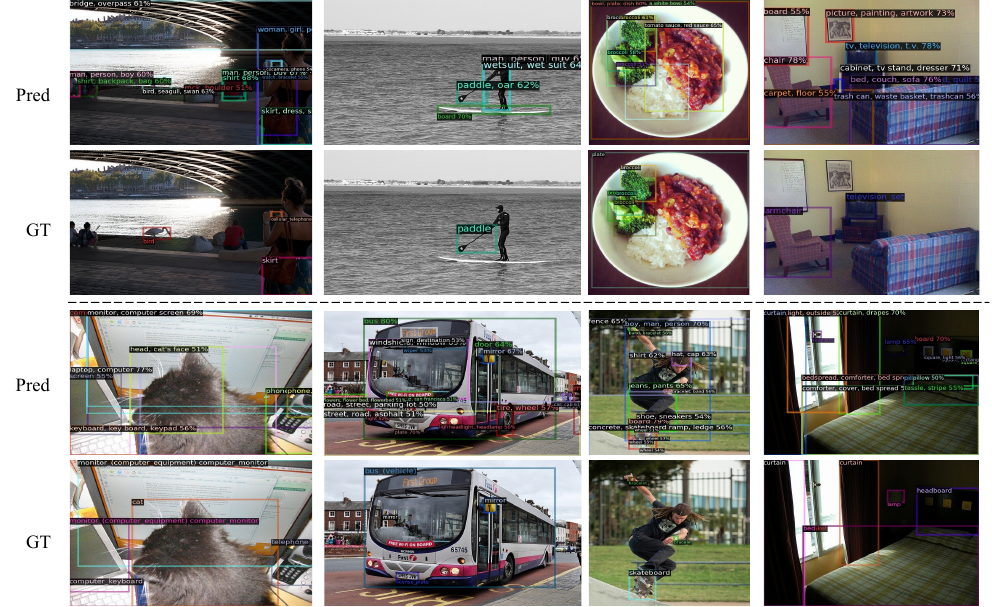}
   \caption{Qualitative prediction results from GenerateU and ground truth on LVIS~\cite{gupta2019lvis}. GenerateU produces complete and precise predictions, showcasing its ability to go beyond fixed vocabulary constraints.}
   \label{fig:lvis_vis}
\end{figure*}

\noindent\textbf{Beam Search.}
Towards a general and practical object detection system,  we believe that beam search plays a crucial role in generating outputs that are closely aligned with human intuition.
This is because it explores a broader array of possibilities during the decoding phase.
Beam search naturally contains the multi-level object names and their synonyms, enhancing the system's ability to recognize objects with varied linguistic expressions.
This contributes to a more adaptable and nuanced understanding of the objects being detected.
As shown in Figure~\ref{fig:beam_search}, beam search largely improves the rare category performance, proving its importance for the long tail problem, and emphasizing the system's effectiveness in handling less frequent object categories.
In our experiments, we set the default beam size to 3.

\noindent\textbf{Evaluation Metrics.}
To evaluate the results of generative object detection, we use different metrics: METEOR, evaluating the quality of text, and similarity scores, calculated by a fixed text encoder.
Note that the off-the-shelf text encoder is not used for the proposed method, but just for the evaluation.
Here we compare two offline models: CLIP text encoder~\cite{radford2021learning} and BERT~\cite{devlin2018bert}, as presented in Table~\ref{tab:metrics}.
We find that the CLIP text encoder is particularly adept at handling noun phrases, while the BERT encoder, predominantly trained on sentence corpora, may exhibit different strengths in processing textual information.
\setlength\tabcolsep{2pt}
\begin{table}
  \begin{tabular}{@{}lclllll@{}}
    \toprule
     \makecell[l]{Pre-Train\\Data} & \makecell[c]{Region-word\\Alignment}& APr & APc & APf & AP \\
    \midrule
    \multirow{2}*{VG} & - & 16.0  & 19.1 & 26.7 & 22.5 \\
    & $\checkmark$ & $17.4_{\uparrow 1.4}$ & $22.4_{\uparrow 3.9}$ & $29.6_{\uparrow 2.9}$ & $25.4_{\uparrow 2.9}$\\
    \midrule
    \multirow{2}*{VG, GRIT} & - & 19.2 & 23.5 & 28.9 & 25.7 \\
    & $\checkmark$ & $20.0_{\uparrow 0.8}$ & $24.9_{\uparrow 1.4}$ & $29.8_{\uparrow 0.9}$ & $26.8_{\uparrow 0.9}$\\
    \bottomrule
  \end{tabular}
  \caption{Effect of region-word alignment loss.}
  \label{tab:alignment_loss}
\end{table}

\noindent\textbf{Region-word Alignment.}
We conduct ablation experiments to evaluate the effectiveness of the region-word alignment loss in our method, as presented in Table~\ref{tab:alignment_loss}. 
We observe that training with only the detection loss and the language modeling loss yields promising results. Adding the region-word alignment loss provides additional supervision and further improves the performance. This further demonstrates the effectiveness of the proposed generative object detection method. 

\subsection{Qualitative Results}
Here we provide qualitative visualizations in showcasing the superior object detection capabilities of our model. 
As shown in Figure~\ref{fig:lvis_vis}, our model can identify a more diverse range of objects compared to ground truth annotations. 
These visualizations not only affirm the reliability of our detection framework but also demonstrate its ability to detect a wider variety of objects, providing valuable insights into the model's strengths.
\section{Conclusion}

This paper introduces GenerateU, a novel approach for generative open-ended object detection. By reformulating object detection as a generative task, the model eliminates the need for predefined categories during inference, offering a more practical solution. Our method achieves comparable results with open-vocabulary models on zero-shot LVIS, even though categories are not seen in our GenerateU. Overall, this work contributes to the fast-evolving area of object detection towards to open-world scenarios. 
Future works include investigating the effect of training data scales and exploring methods beyond naive pseudo-labeling.
{
    \small
    \bibliographystyle{ieeenat_fullname}
    \bibliography{main}
}
\clearpage
\thispagestyle{empty} 
\twocolumn[
\begin{@twocolumnfalse}
\begin{center}
\vspace*{-20pt} 
\textbf{\Large Supplementary Material}
\end{center}
\end{@twocolumnfalse}
]
\setcounter{section}{0}
\section{Training loss}
In this section, we elaborate on the training loss mechanism employed in our model.
\subsection{Region-language alignment loss}
\begin{figure}[H]
  \centering
    \includegraphics[width=1.0\linewidth]{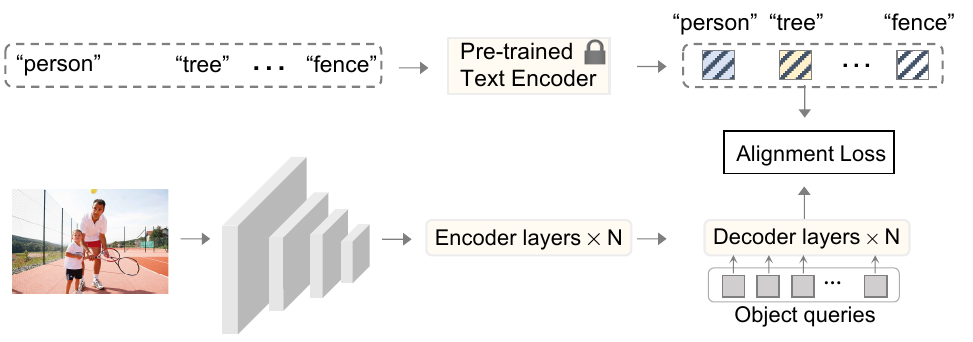}
    \caption{Region-language alignment loss. This process involves aligning text features extracted from a pre-trained CLIP text encoder with object queries of each layer of the decoder. This alignment is for enhancing the representation of visual features.}
    \label{fig:align}
\end{figure}
In the original DETR model, auxiliary losses are incorporated to predict categories and localize results in each of the decoder layers during training, which then contribute to the deep supervision.
However, in our open-world object detector approach, we modify this strategy by integrating a fixed clip text encoder, which replaces the additional supervision in the model training, see Figure~\ref{fig:align}.
At the output of each of the six decoder layers, we align the region feature with the text embedding from this fixed clip text encoder. For each object query, its corresponding word in the text encoder is treated as a positive sample, while other words in the same minibatch are considered negative samples $W'$. We apply binary cross-entropy loss for this process:

\begin{equation}
L_{align} = \sum_{i=1}^{M}-\left[\log\sigma(s_{ik}) + \sum_{t \in W' }log(1-\sigma(s_{it}))\right],
\label{region-word_v1}
\end{equation}
where $\sigma$ is the sigmoid activation, $s_{ik}$ is the alignment score of the $i$-th region feature and its corresponding $k$-th word embedding.

\subsection{Language modelling loss}
The object detector extracts features for each detected object in the image $I$, and these features $f_i^I$ are then used as the context for generating text.
We use cross-entropy loss for each word in the generated text.
$\{f_i\}_{i=1,...,M}$ represents the set of features extracted from the objects detected in the image.
$P(w_i^t | f_i, w_i^1, ..., w_i^{t-1})$ is the probability of the next word $w_t$ given the object features and the preceding words in the text sequence.
The language modelling loss is computed as
\begin{equation}
    L_{lm} = \sum_{i=1}^{M}(-\sum_{t=1}^{N} \log P(w_i^t | f_i, w_i^1, ..., w_i^{t-1})).
  \label{eq:important}
\end{equation}
The goal is to minimize the negative log likelihood of the correct word predictions, which encourages the model to generate accurate and relevant text based on the object features.

Besides, the open-world object detector employs binary cross-entropy for classifying the foreground and background $L_{bce}$, L1 regression loss for box localization $L_{l1}$, and generalized IoU for improved accuracy $L_{gIoU}$.
Finally, our overall loss function is given by
\begin{equation}
    L = L_{bce} + L_{l1} + L_{gIoU} + L_{lm} + L_{align}
  \label{eq:important}
\end{equation}

\begin{figure}
  \centering
    \includegraphics[width=1.0\linewidth]{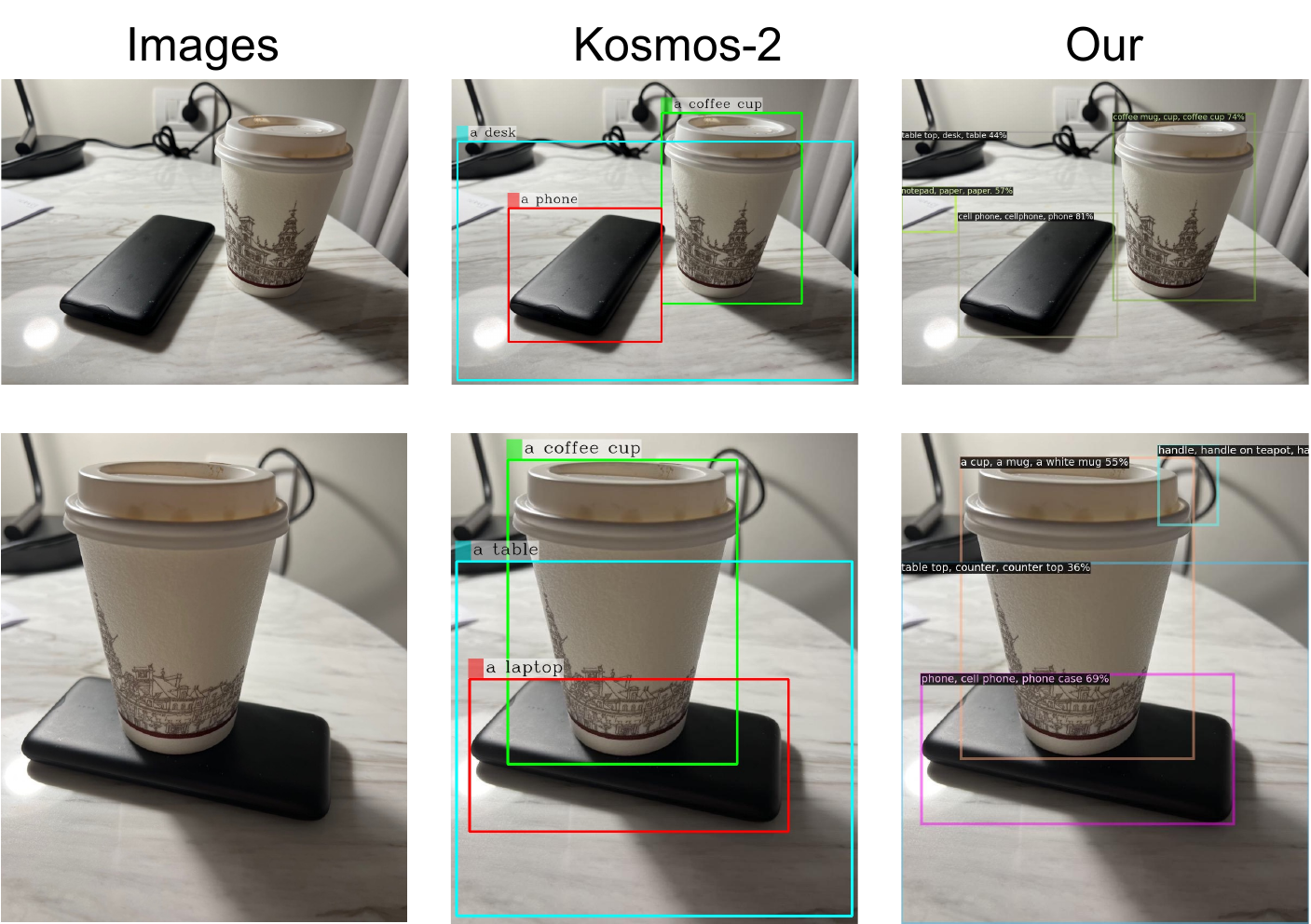}
    \caption{Compared with Kosmos-2. Notably, when the positions of the ``phone" and ``cup" were altered, Kosmos-2 yielded incorrect results, misidentifying the objects. In contrast, our model, despite its simpler architecture, maintained precise and consistent object identification. This highlights the robustness of our approach.}
    \label{fig:supp}
\end{figure}

\section{Compared with MLLM}

Recently, advancements in Multimodal Large Language Models (MLLMs) have enabled the capability to localize objects in images based on generated text descriptions, as discussed in existing literature Kosmos-2~\cite{peng2023kosmos} and DetGPT~\cite{pi2023detgpt}. 
Such MLLMs, however, are not equipped for dense prediction tasks, rendering direct comparisons using detection mean Average Precision (mAP) metrics infeasible. 
To demonstrate the superiority of our model in contrast to Kosmos-2, we present several case studies. 
One notable example is illustrated in Figure~\ref{fig:supp}, where the images contain two common objects: a mobile phone and a cup. 
When these items were placed separately, both models accurately identified them. 
The distinction in performance became evident when the objects' arrangement was altered, specifically with the cup positioned atop the phone. 
In this setup, Kosmos-2, heavily reliant on its language model, erroneously classified the phone as a laptop, highlighting its vulnerability to context-based errors. 
In contrast, our model exhibited remarkable consistency and robustness, correctly identifying the objects regardless of their configuration. This test underscores our model's enhanced reliability in complex, real-world environments, emphasizing its suitability for a wide range of object recognition applications.
\section{Pseudo-labeling Method Details}

Our pseudo-labeling method is a two-step process. 
\textbf{1) \textit{Initial labels from KOSMOS-2.}} We directly utilize the dataset of GRIT image-text pairs [37], where the images are sourced from COYO700M [5]. For these images, the initial set of pseudo-grounding labels is from KOSMOS-2. However, we found that the count of nouns in the captions is generally much lower than the actual number of objects in the images, which suggests that such initial pseudo-labels are not comprehensive enough to cover all/most objects in the images.
\textbf{2)  \textit{Supplement with GenerateU.}} To address the limitation, we incorporated GenerateU, trained on the Visual Genome (VG) dataset, to supplement additional objects in the images that may not be covered by the initial pseudo-labels from KOSMOS-2. We use post-processing (e.g., NMS) and a confidence score greater than 0.5 to filter the results. We generate the pseudo-labels only once. In Figure 3, the pseudo-labels with a black underline are from KOSMOS-2, and the others are generated by GenerateU. This step is crucial for enhancing the comprehensiveness of the pseudo-label generation process.
The table below shows the effectiveness of GenerateU's supplemental labels. 
\begin{table}[H]
\begin{center}
\setlength{\tabcolsep}{1.2mm}{
\begin{tabular}{l|cccc}
\hline
Zero-shot LVIS & APr & APc & APf & AP\\
\hline
Initial labels  & 19.3 & 21.8 & 29.0 & 25.0\\
\hline
Supplement with GenerateU & 20.0 & 24.9 & 29.8 & 26.8 \\
\hline
\end{tabular}
}
\end{center}
\caption{The effectiveness of GenerateU's supplemental labels.}
\end{table}

\balance
   

\end{document}